\documentclass[10pt,twocolumn,letterpaper]{article}
\usepackage{spconf,amsmath,graphicx}
\usepackage{mdwmath}
\usepackage{mdwtab}
\usepackage{url}
\usepackage{CJK}
\usepackage{amsfonts}
\usepackage{amssymb}

\newcommand\eightpt{\def\baselinestretch{0.95}\let\normalsize\normalsize\scriptsize}
\newcommand\sevenpt{\def\baselinestretch{1.5}\let\normalsize\normalsize\tiny}
\newcommand{\zhs}[1]{\begin{CJK}{UTF8}{gbsn}#1\end{CJK}}

\title{Embedded Large--Scale Handwritten Chinese Character Recognition}

\name{\sl Youssouf Chherawala, Hans J.G.A. Dolfing, Ryan S. Dixon, {\rm and} Jerome R. Bellegarda}

\address{Apple Inc., Cupertino, California 95014, USA}

\begin{document}
\maketitle
\baselineskip 10pt

\begin{abstract}
  As handwriting input becomes more prevalent, the large symbol
  inventory required to support Chinese handwriting recognition poses
  unique challenges. This paper describes how the Apple deep learning
  recognition system can accurately handle up to 30,000 Chinese
  characters while running in real-time across a range of mobile
  devices.  To achieve acceptable accuracy, we paid particular
  attention to data collection conditions, representativeness of
  writing styles, and training regimen. We found that, with proper
  care, even larger inventories are within reach. Our experiments show
  that accuracy only degrades slowly as the inventory increases, as
  long as we use training data of sufficient quality and in sufficient
  quantity.
\end{abstract}


\zhs{} 

\section{Introduction}
Online handwriting input has recently become more prevalent given the
pervasiveness of mobile phones, tablets, and wearable gear like
smartwatches. For Chinese in particular, it can significantly enhance
user experience given the relative complexity of keyboard
methods. Chinese handwriting recognition is uniquely challenging, due
to the large number of distinct entries in the underlying character
inventory. Unlike alphabet-based writing, which typically involves on
the order of 100 symbols, the standard set of H\`anz\`i characters in
Chinese National Standard Gu\'oji\=a Bi\=aozh\v un GB18030--2005
contains 27,533 entries, and many additional logographic characters
are in use throughout Greater China.

For computational tractability, it is usual to focus on a restricted
number of ``commonly used'' characters. The standard GB2312-80 set
only includes 3,755 (level-1) and 3,008 (level-2) entries, for a total
of 6,763 characters. The closely aligned character set used in the
popular CASIA databases comprises a total of 7,356 entries
\cite{casia}. The handwritten database SCUT-COUCH has similar coverage
\cite{scutcouch}.  At the individual user level, however, what is
``commonly used'' typically varies somewhat from one person to the
next. Most people need at least a handful of characters deemed
``infrequently written,'' as they happen to occur in the compendium of
proper names that is relevant to them. Thus ideally we would need to
scale up to at least the level of GB18030-2005.

While early recognition algorithms mainly relied on structural methods
based on individual stroke analysis, the need to achieve stroke-order
independence later sparked interest into statistical methods using
holistic shape information \cite{pami2004}. This obviously complicates
large-inventory recognition, as correct character classification tends
to get harder with the number of categories to disambiguate
\cite{hinton85}. On Latin script tasks such as MNIST \cite{mnist},
convolutional neural networks (CNNs) soon emerged as the approach of
choice \cite{simard}. Given a sufficient amount of training data,
supplemented with synthesized samples as necessary, CNNs decidedly
achieved state-of-the-art results \cite{ciresan, meier}. The number of
categories in those studies, however, was by nature very small (10).

Applying CNNs to the large-scale recognition of Chinese characters
thus requires careful consideration of the underlying character
inventory. This is particularly true if the recognition system is to
perform inference in real-time on mobile devices. This paper focuses
on the challenges involved in terms of accuracy, character coverage,
and robustness to writing styles. We investigate and adapt recent
advances in CNN architecture design for the unique challenges of large
datasets, in order to produce practical models which can be deployed
in commercial applications. Section~\ref{sec:cnn} presents the CNN
architecture we adopted. Section~\ref{sec:dataset} focuses on the
challenges involved in scaling up the system to a larger
inventory. Section~\ref{sec:experiments} describes our experiments and
presents comparative results obtained on the CASIA database. Finally,
Section~\ref{sec:discussion} concludes with some prognostications regarding
a possible evolution to the full Unicode inventory.

\section{Model Architecture}
\label{sec:cnn}
We adopt the MobileNetV2 CNN architecture~\cite{mobilenetV2} and adapt
it to our Chinese handwriting recognition task, due to: (i) impressive
accuracies observed for image classification, and (ii) real-time
performance on mobile devices. The first aspect can be traced to both
residual connections~\cite{residualConnections}, which allow training
of deeper and more accurate networks, and batch
normalization~\cite{batchNormalization}, which enables faster model
convergence during training while acting as a regularizer.  The second
aspect involves replacing standard convolutions with depthwise
separable convolutions~\cite{depthwiseSeparable}, which reduce by an
order of magnitude the number of operations needed. In addition,
MobileNetV2 relies on the so-called (inverted) {\it bottleneck block},
which further reduces memory requirements at inference time. The
number of channels of the input of the block is first increased by an
expansion factor $t$ before performing the convolution and compressed
back at the output of the block.  Maintaining the compressed form of
the input in memory for the residual connection reduces the memory
footprint.

\begin{figure}[t]
\centering
\includegraphics[clip, trim=0.6cm 10.0cm 0.6cm 0.5cm, width=3.5in]{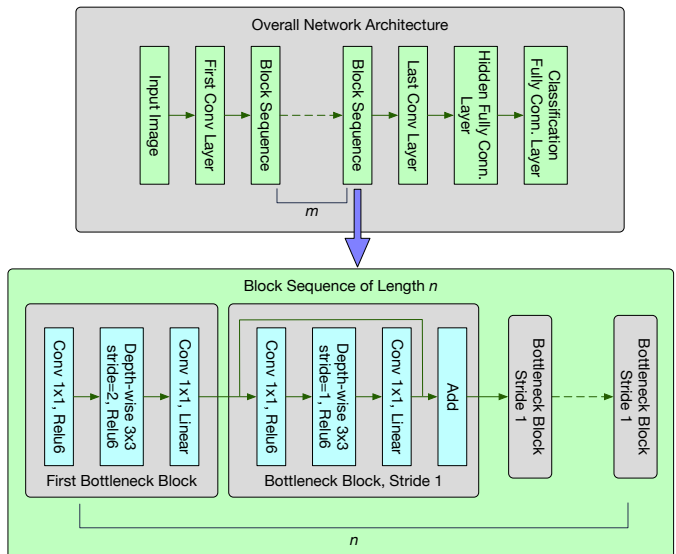} 
\vspace{-10truemm}
\caption{\sl Adapted MobileNetV2 Architecture.}
\label{fig:mobileNetV2}
\vspace{-2truemm}
\end{figure}

Our adapted version of MobileNetV2 is shown in
Fig.~\ref{fig:mobileNetV2}, with the \textit{block sequence} as main
building block. Each such block is itself a sequence of $n$ bottleneck
blocks. Only the first bottleneck block of each sequence performs
spatial subsampling by using a stride of 2. The remaining bottleneck
blocks have residual connections between their input and
outputs. Different bottleneck blocks can be sequences of different
lengths. The input of the network is a medium-resolution image (for
performance reasons) of $48\times 48$ pixels representing a Chinese
handwritten character. It is fed to a convolution layer, followed by
$m$ block sequences, a last convolution layer, and a fully-connected
layer before the classification layer. The first and last convolution
layer do not perform spatial subsampling. Therefore, spatial
subsampling is solely controlled by the number of block
sequences. Finally, the classification layer has one node per class,
e.g., 3,755 for the H\`anz\`i level-1 subset of GB2312-80, and close
to 30,000 when scaled up to the full inventory (cf.  Section~3).

At that scale, disk footprint is dominated by the parameters of the
fully-connected classification layer. It is very important to control
the input dimension of that last layer in order to avoid generating
models with a very large number of parameters, which would not only be
hard to train but would also have an absurdly large footprint. This
can be avoided by using a very conservative number of output units in
the last hidden layer. The same reasoning applies as well to the input
of the hidden fully-connected layer. Its input dimension can be
restricted by stacking a sufficient number of block sequences to
decrease the image spatial resolution and by setting a conservative
number of channels for the last convolution layer.

Note that, given our product focus, we deliberately do not tune our
system for the highest possible accuracy on benchmark datasets. Indeed
our priorities are model size, evaluation speed, and user
experience. In that context, we opt for a compact system that works in
real-time, across a wide variety of styles, and with high robustness
towards non-standard stroke order. This leads to an image based
recognition approach even though we evaluate it on on-line datasets.

\section{Scaling Up to 30K Characters}
\label{sec:dataset}
Since the ideal set of ``frequently written'' characters varies from
one user to the next, a large population of users requires an
inventory of characters much larger than 3,755. Exactly which ones to
select, however, is not entirely straightforward. Simplified Chinese
characters defined with GB2312-80 and traditional Chinese characters
defined with Big5, Big5E, and CNS 11643-92 cover a wide range (from
3,755 to 48,027 H\`anz\`i characters). More recently came HKSCS-2008
with 4,568 extra characters, and even more with GB18030-2000.

\begin{figure}[t]
\centering
\fbox{\includegraphics[width=0.7in,height=0.7in]{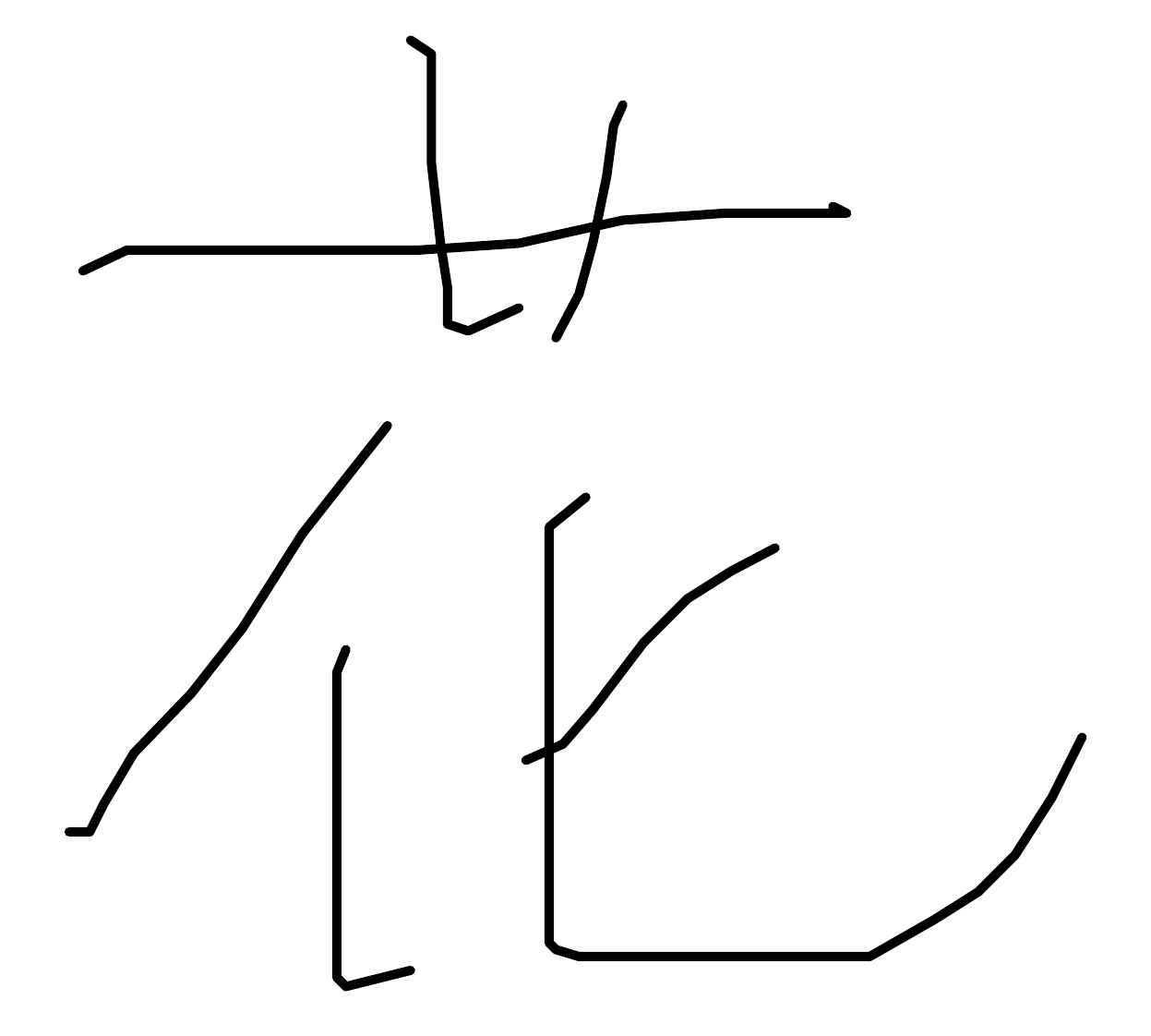}}
\fbox{\includegraphics[width=0.7in,height=0.7in]{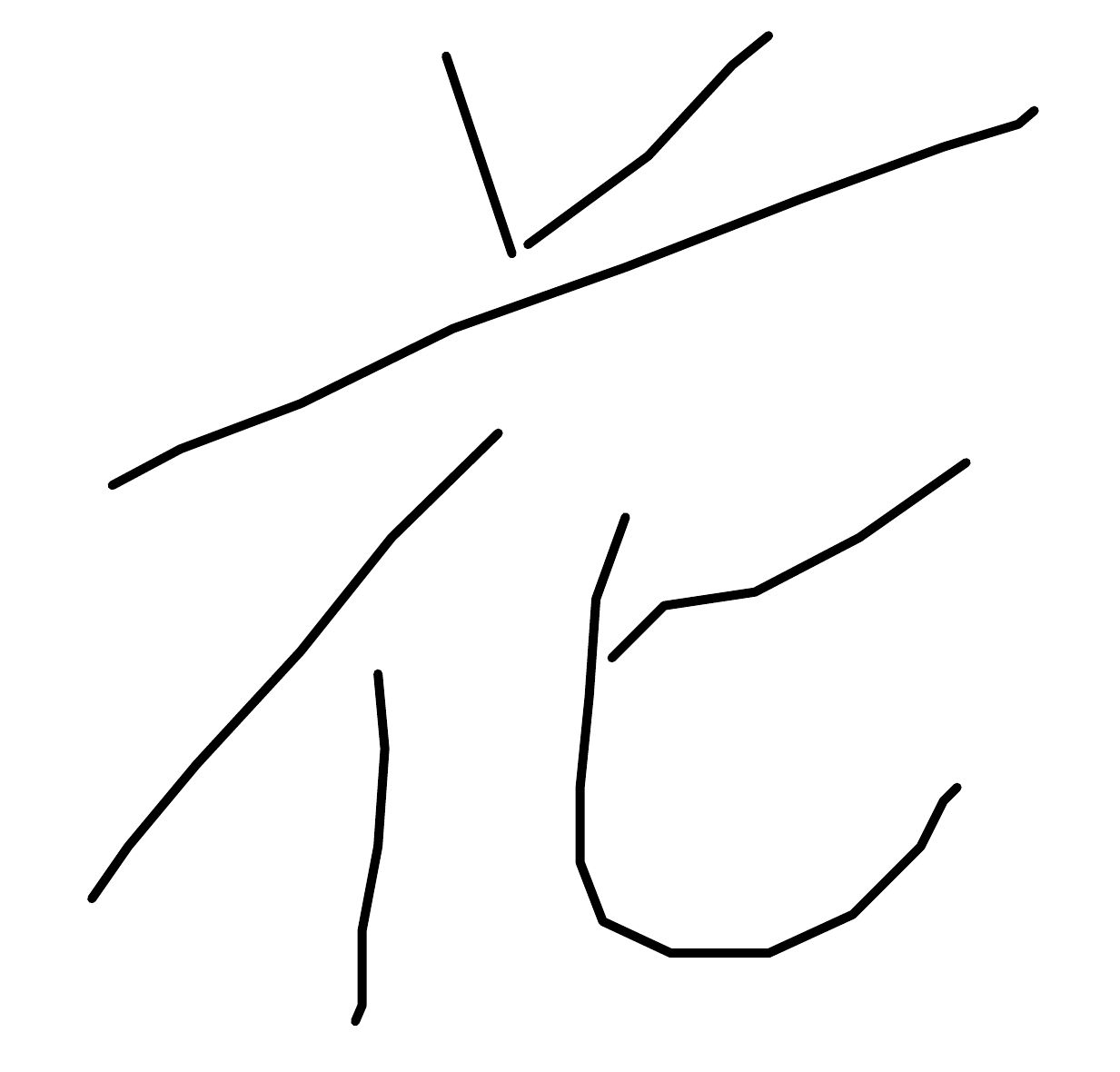}}
\fbox{\includegraphics[width=0.7in,height=0.7in]{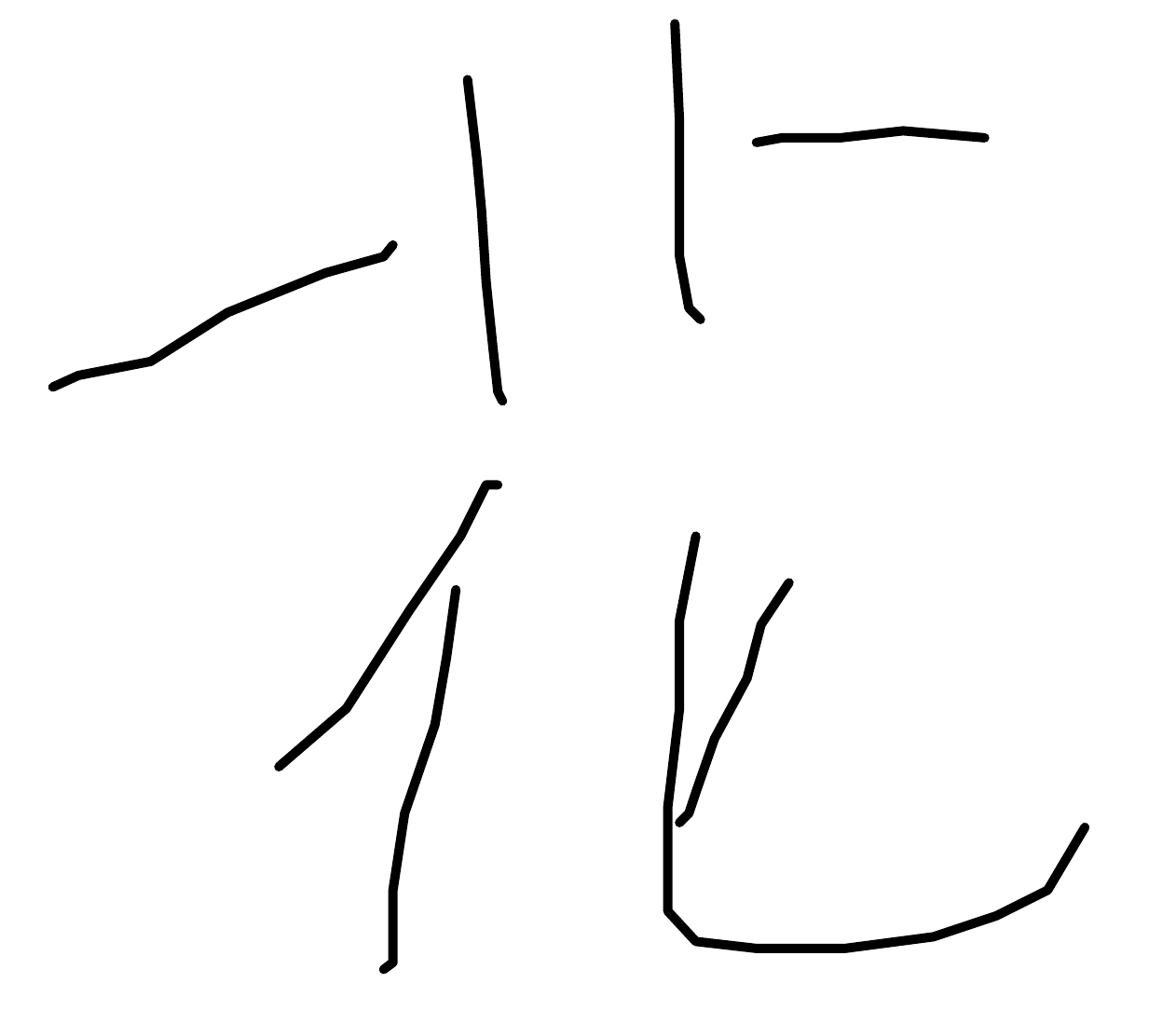}}
\fbox{\includegraphics[width=0.7in,height=0.7in]{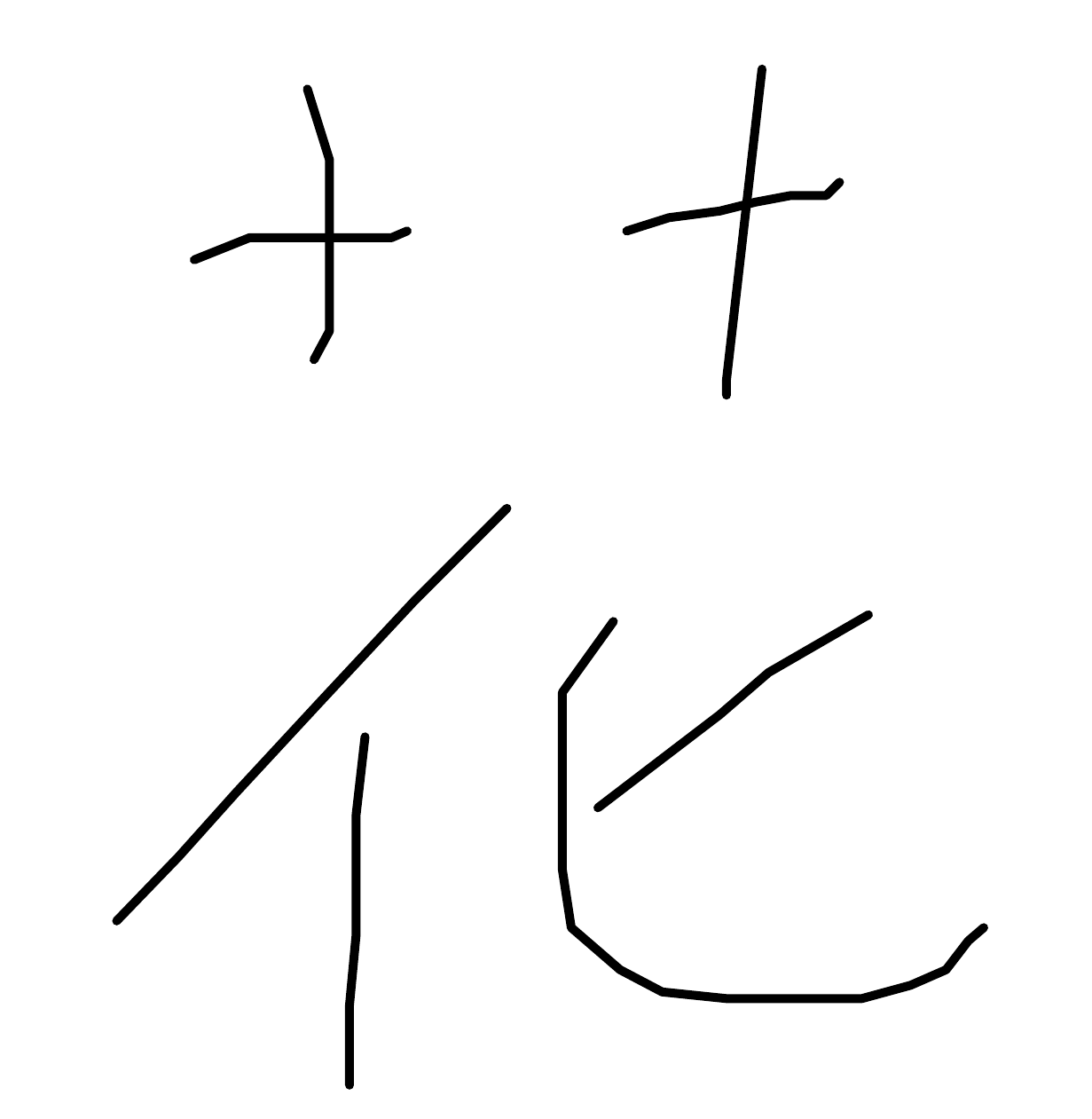}}
\vspace{-3truemm}
\caption{\sl Printed Radical Variations of U+82B1 (\protect\zhs{花}) } 
\label{fig1}
\end{figure}

\begin{figure}[t]
\centering
\vspace{-2truemm}
\fbox{\includegraphics[width=0.7in,height=0.7in]{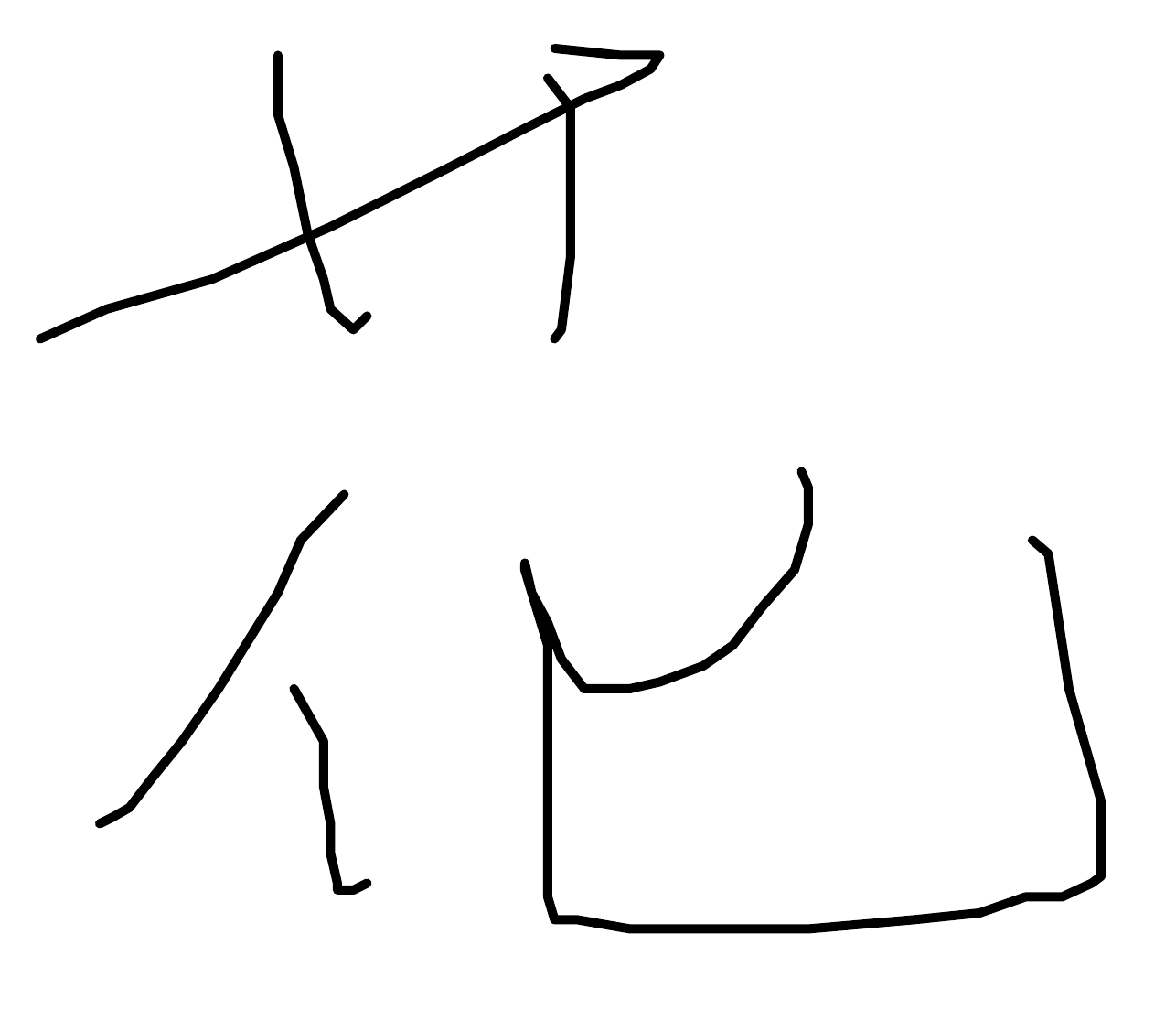}}
\fbox{\includegraphics[width=0.7in,height=0.7in]{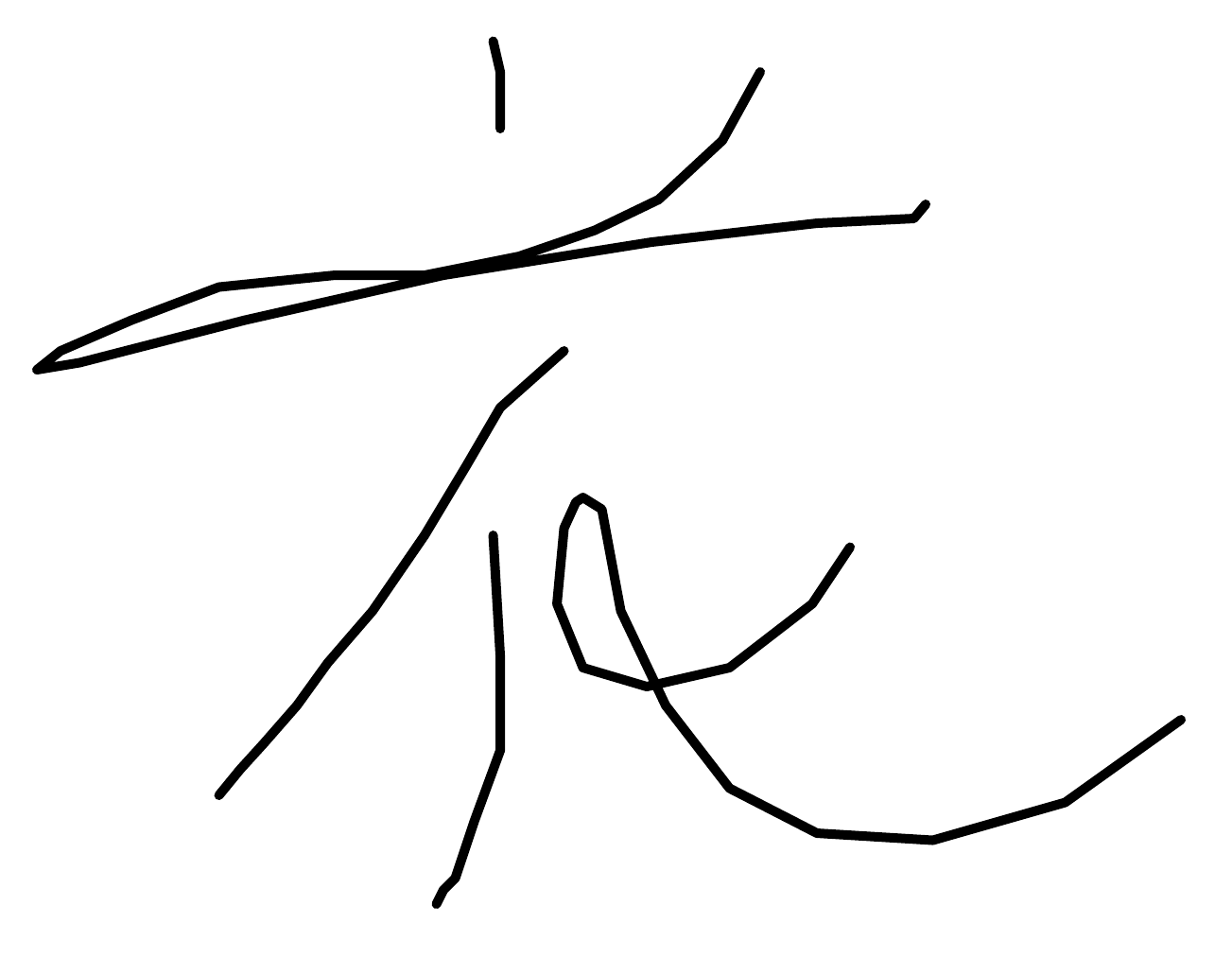}}
\fbox{\includegraphics[width=0.7in,height=0.7in]{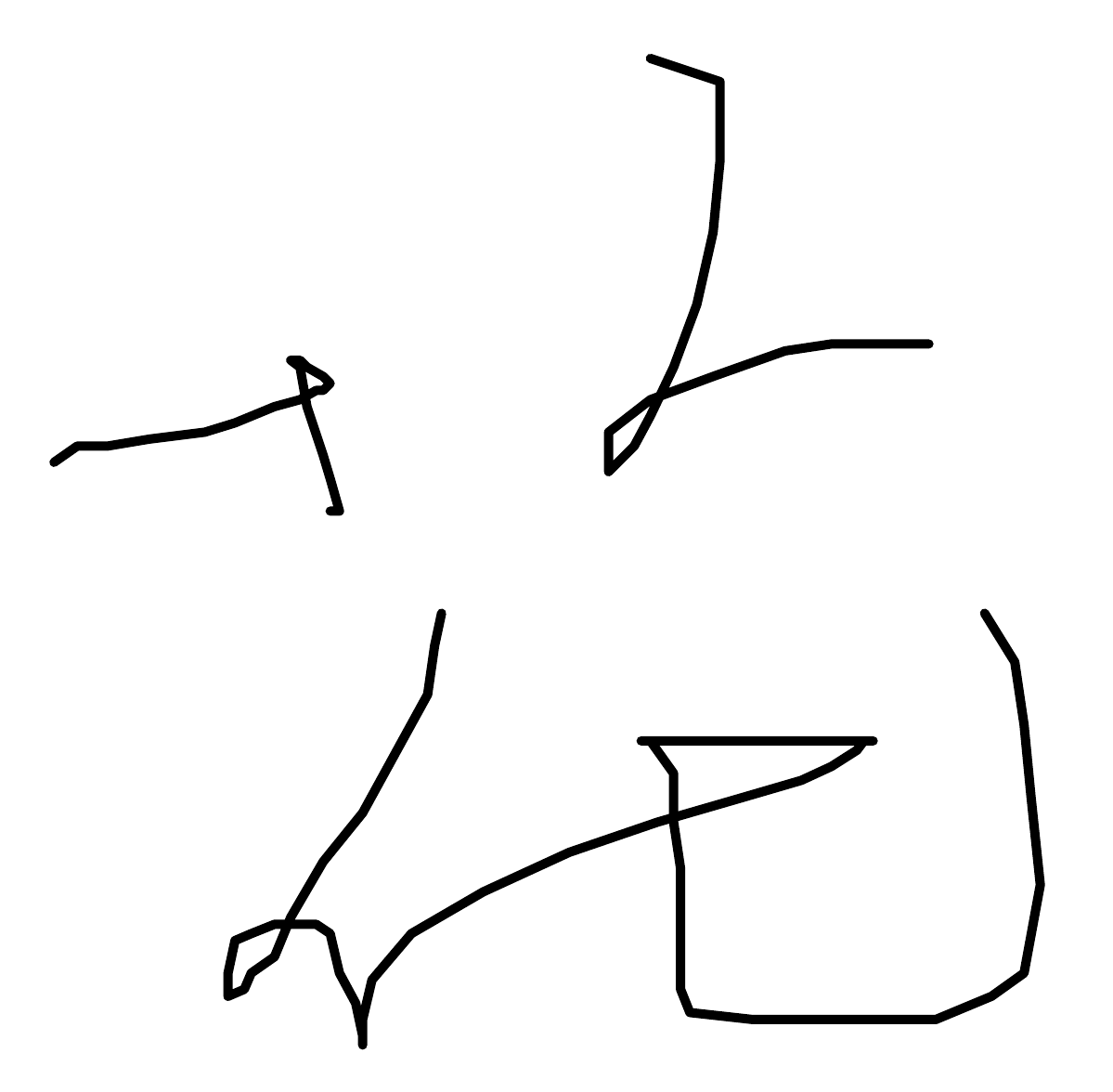}}
\fbox{\includegraphics[width=0.7in,height=0.7in]{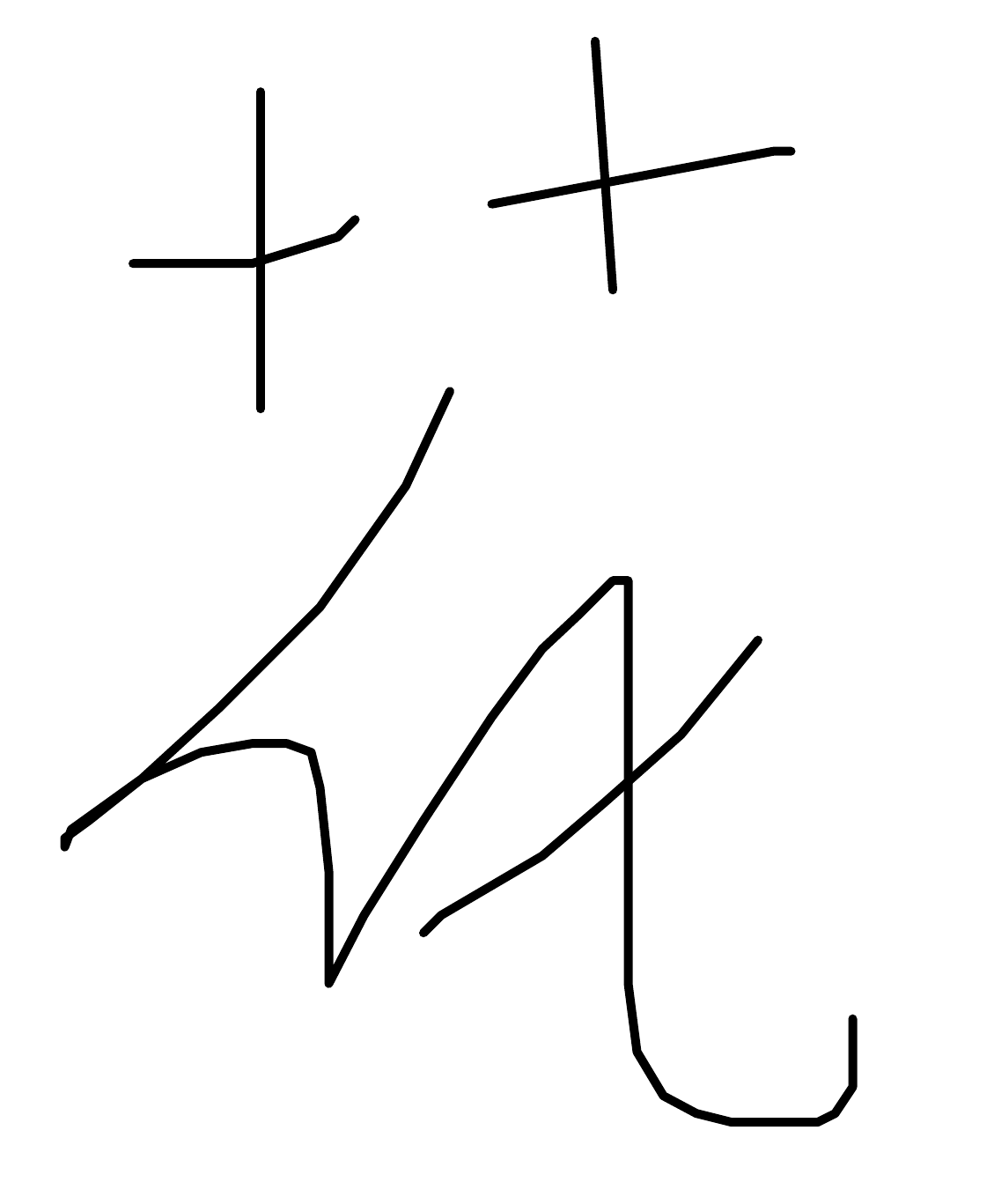}}
\vspace{-3truemm}
\caption{\sl Cursive Radical Variations of U+82B1 (\protect\zhs{花}) }
\label{fig2}
\end{figure}

\begin{figure}[!ht]
\vspace{-2truemm}
\centering
\fbox{\includegraphics[width=0.7in,height=0.7in]{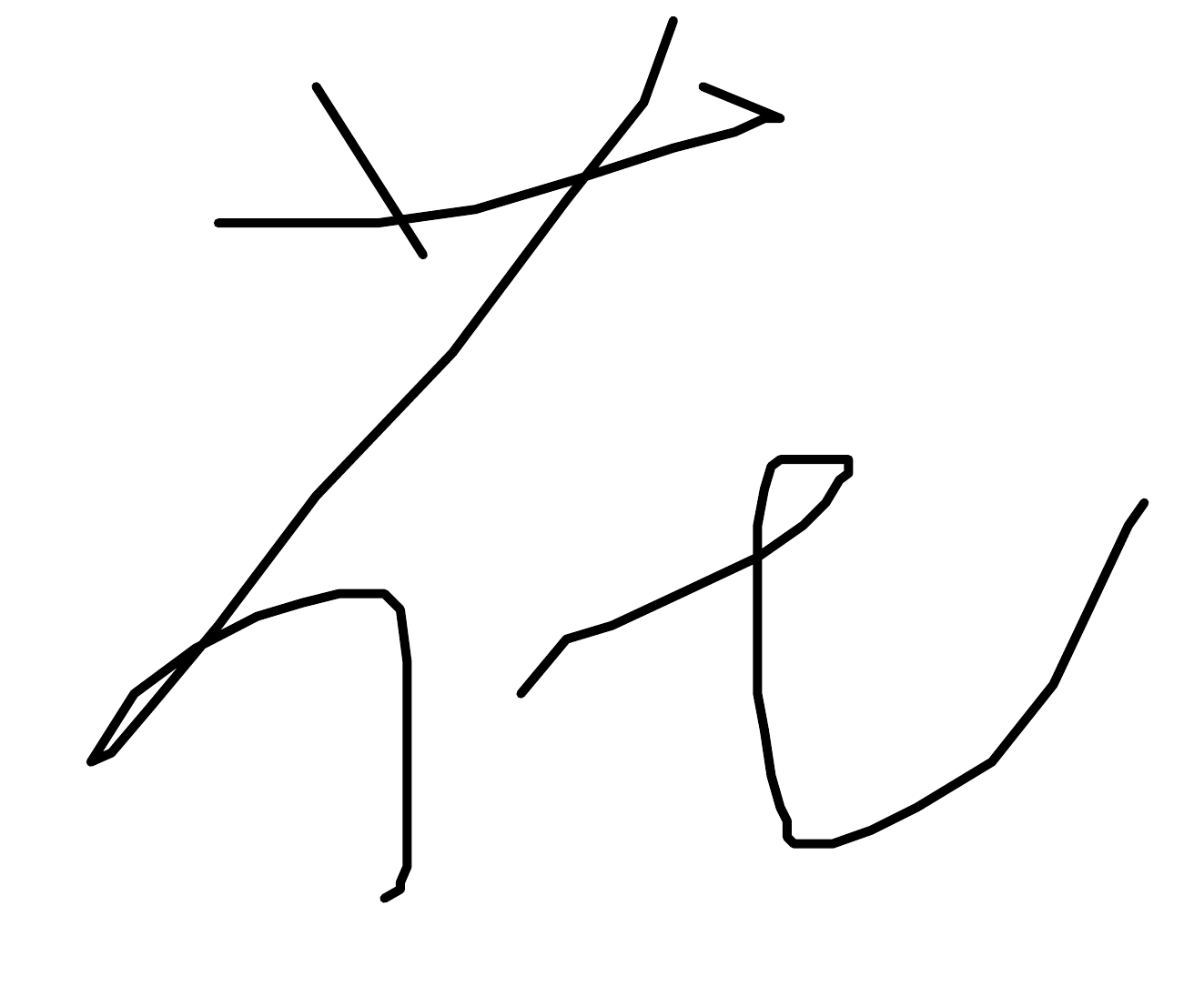}}\hspace{5truemm}
\fbox{\includegraphics[width=0.7in,height=0.7in]{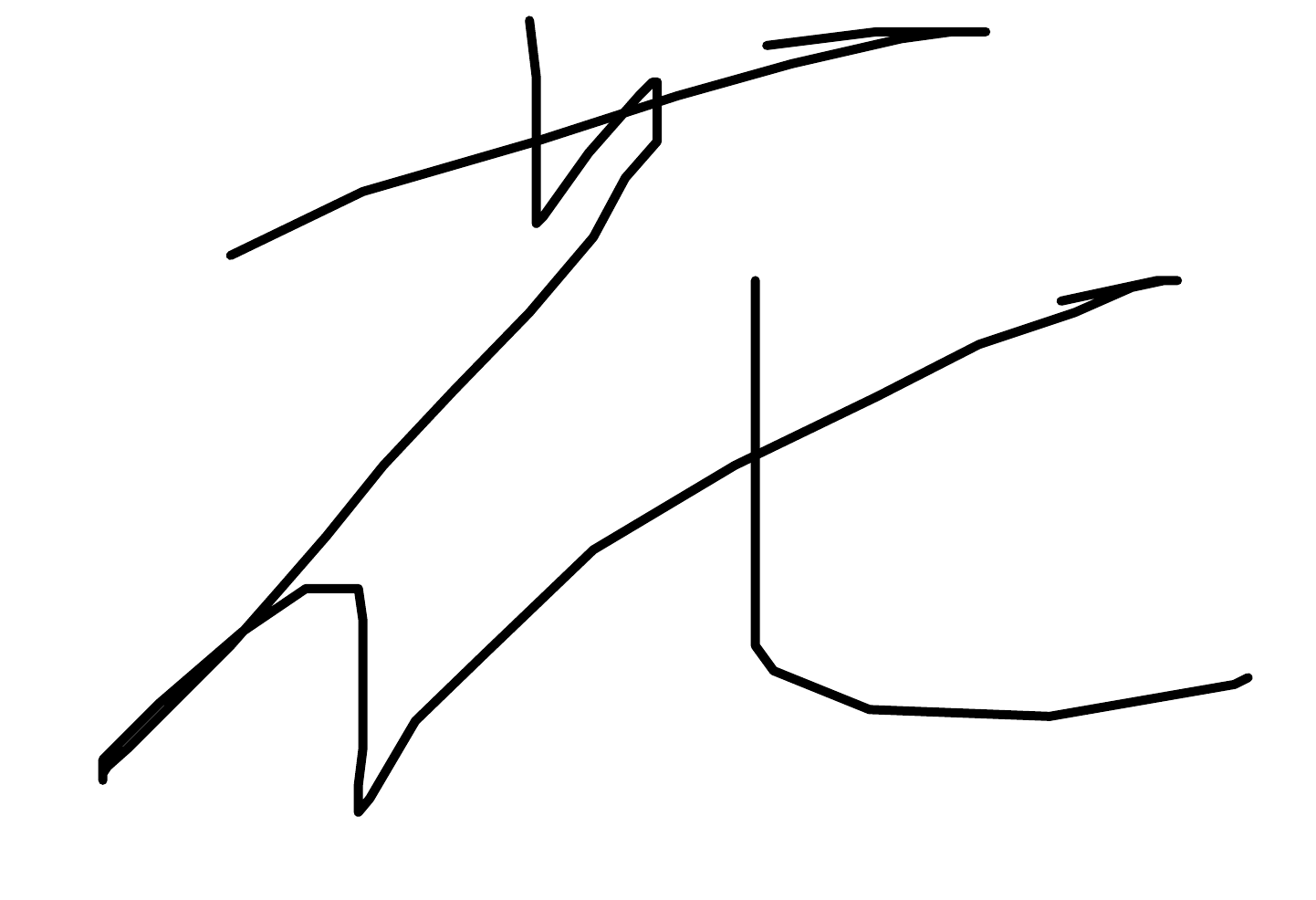}}\hspace{5truemm}
\fbox{\includegraphics[width=0.7in,height=0.7in]{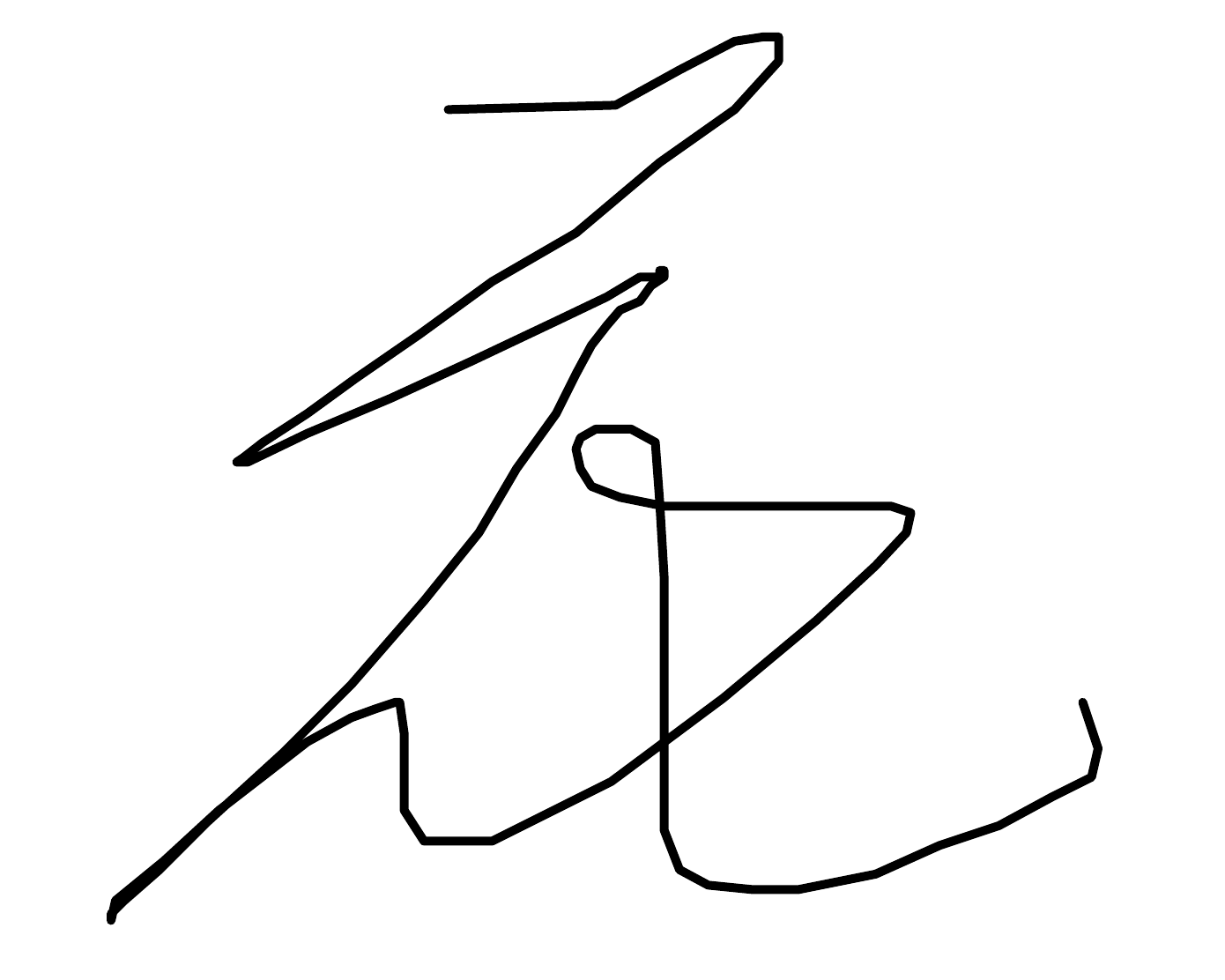}}
\vspace{-3truemm}
\caption{\sl Unconstrained Variations of U+82B1  (\protect\zhs{花})  }
\label{fig3}
\vspace{-2truemm}
\end{figure}

We opted for the H\`anz\`i part of GB18030-2005, HKSCS-2008, Big5E, a
core ASCII set, and a set of visual symbols and emojis, for a total of
approximately 30,000 characters, which we felt represented the best
compromise for the daily correspondence of most Chinese users.

 Upon selection of the character inventory, it is critical to sample
the writing styles that users actually use. While there are formal
clues as to what styles to expect (cf. \cite{cursive}), there exist
regional variations in writing styles, e.g., (i) the use of the U+2EBF
(\zhs{艹}) radical, or (ii) the cursive U+56DB (\zhs{四}) vs. U+306E
(\zhs{の}). Rendered fonts can also contribute to confusion as some
users expect specific characters to be presented in a specific style.
As a speedy input tends to drive toward cursive styles, it tends to
increase ambiguity, e.g. between U+738B (\zhs{王}) and U+4E94
(\zhs{五}).  Finally, increased internationalization sometimes
introduces unexpected collisions: for example, a cursively written
U+4E8C (\zhs{二}) may conflict with the Latin characters ``2'' and
``Z''.

To cover the whole spectrum of possible input from printed to cursive
to unconstrained writing \cite{pami2004}, including as many variants
as we could, we sought data from a broad spectrum of writers from
multiple regions in Greater China.  We collected data from paid
participants across various age groups, gender, and with a variety of
educational backgrounds, thus resulting in tens of millions of
character instances for our training data.  We were surprised to
observe that most users have never seen, let alone written, many of
the rarer characters. This unfamiliarity results in hesitations,
stroke-order errors, and other distortions which introduce additional
complexity, as the data collection needs to capture such distortions
in sufficient detail that models can properly encapsulate them.

To illustrate the variety of writing styles we collected,
Figs.~\ref{fig1}--\ref{fig3} show some examples of the ``flower''
character for U+82B1 (\zhs{花}).  Fig.~\ref{fig1} illustrates the
variety in the grass radical on the top, Fig.~\ref{fig2} does the same
for cursive, and Fig.~\ref{fig3} for even more unconstrained data.

\begin{figure}[h]
\centering
\fbox{\includegraphics[width=0.7in,height=0.7in]{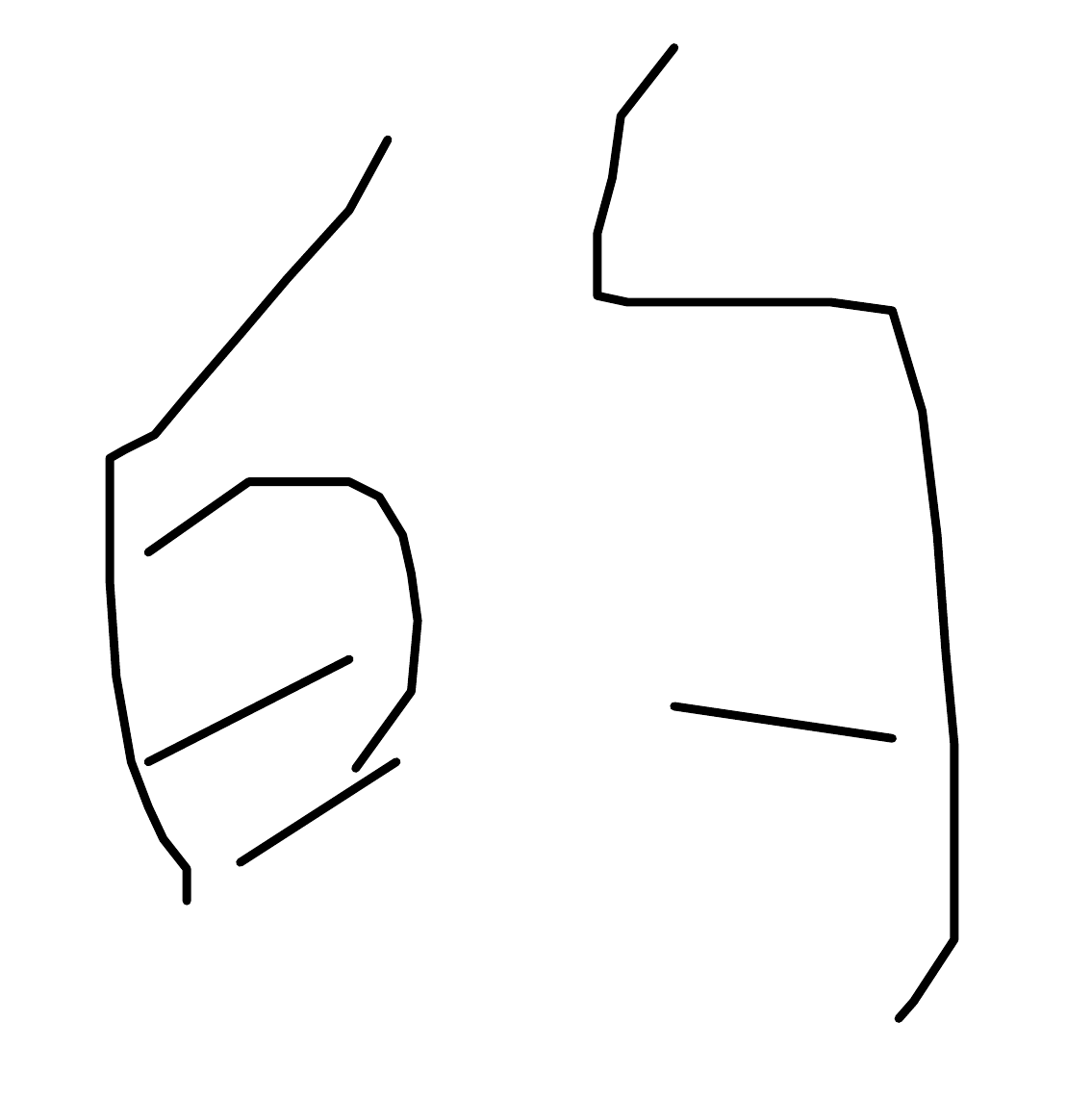}}\hspace{5truemm}
\fbox{\includegraphics[width=0.7in,height=0.7in]{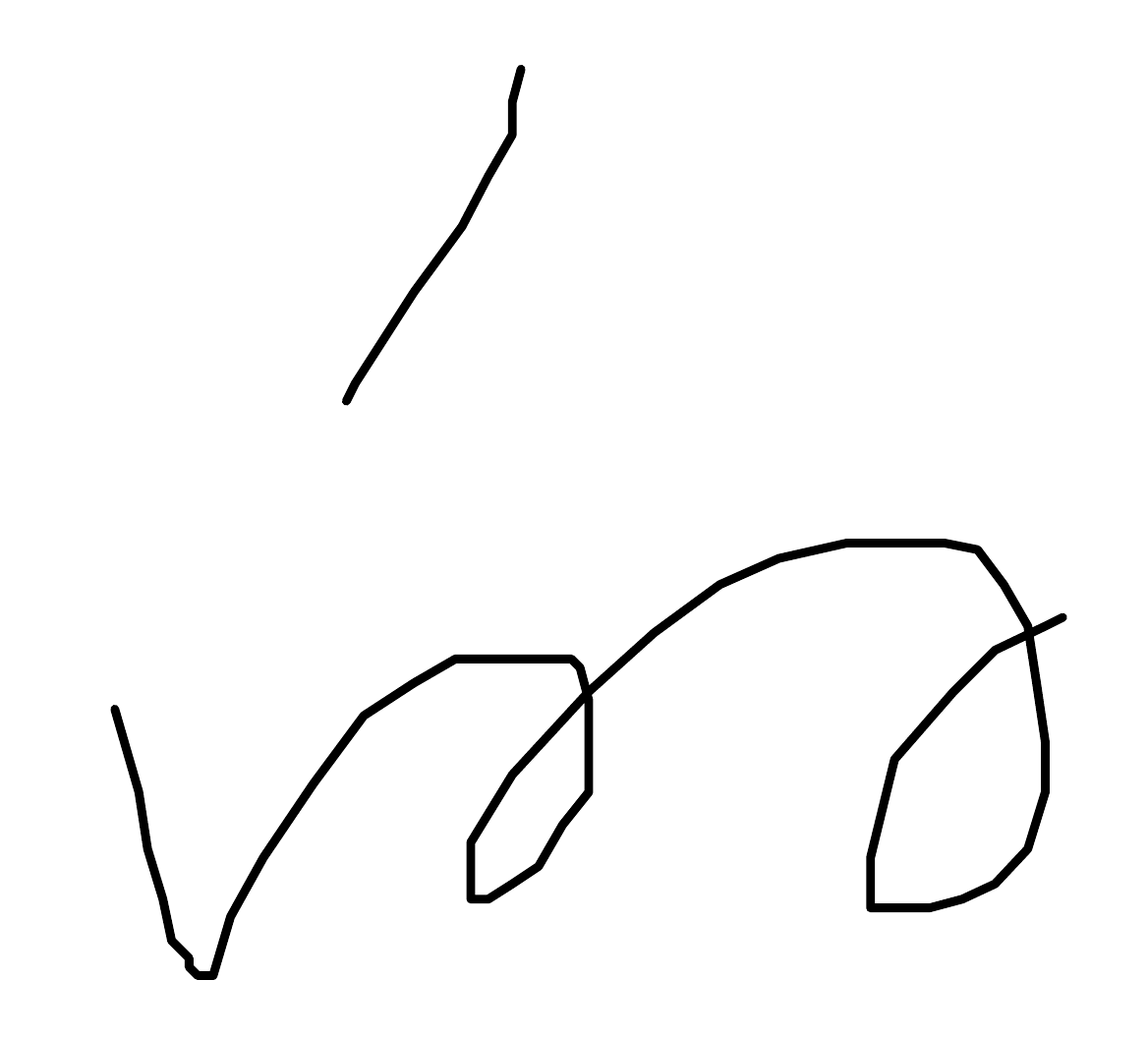}}\hspace{5truemm}
\fbox{\includegraphics[width=0.7in,height=0.7in]{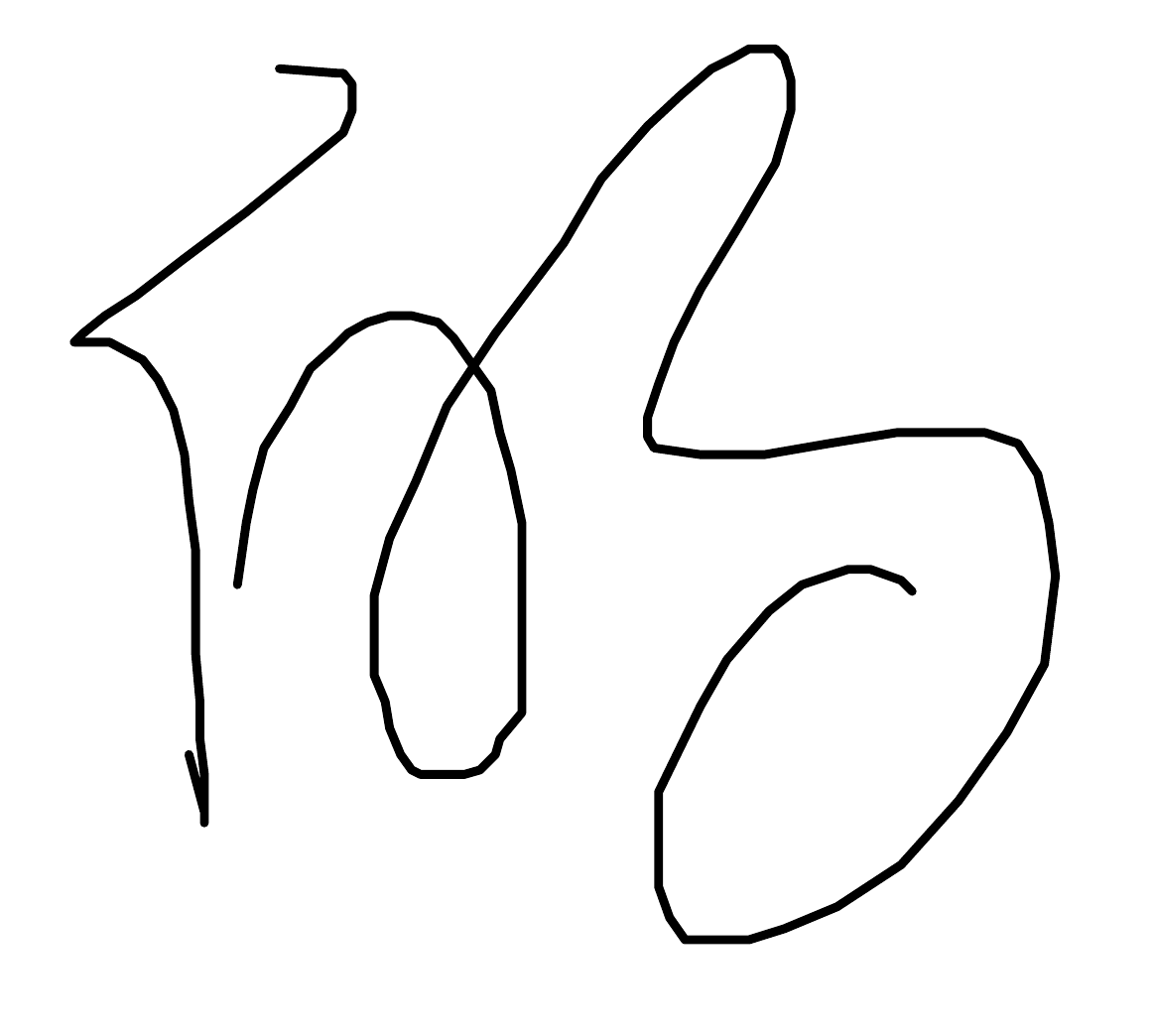}}
\vspace{-3truemm}
\caption{\sl Variations of U+7684  (\protect\zhs{的})  }
\label{duprint}
\end{figure}

\begin{figure}[!h]
\vspace{-2truemm}
\centering
\fbox{\includegraphics[width=0.7in,height=0.7in]{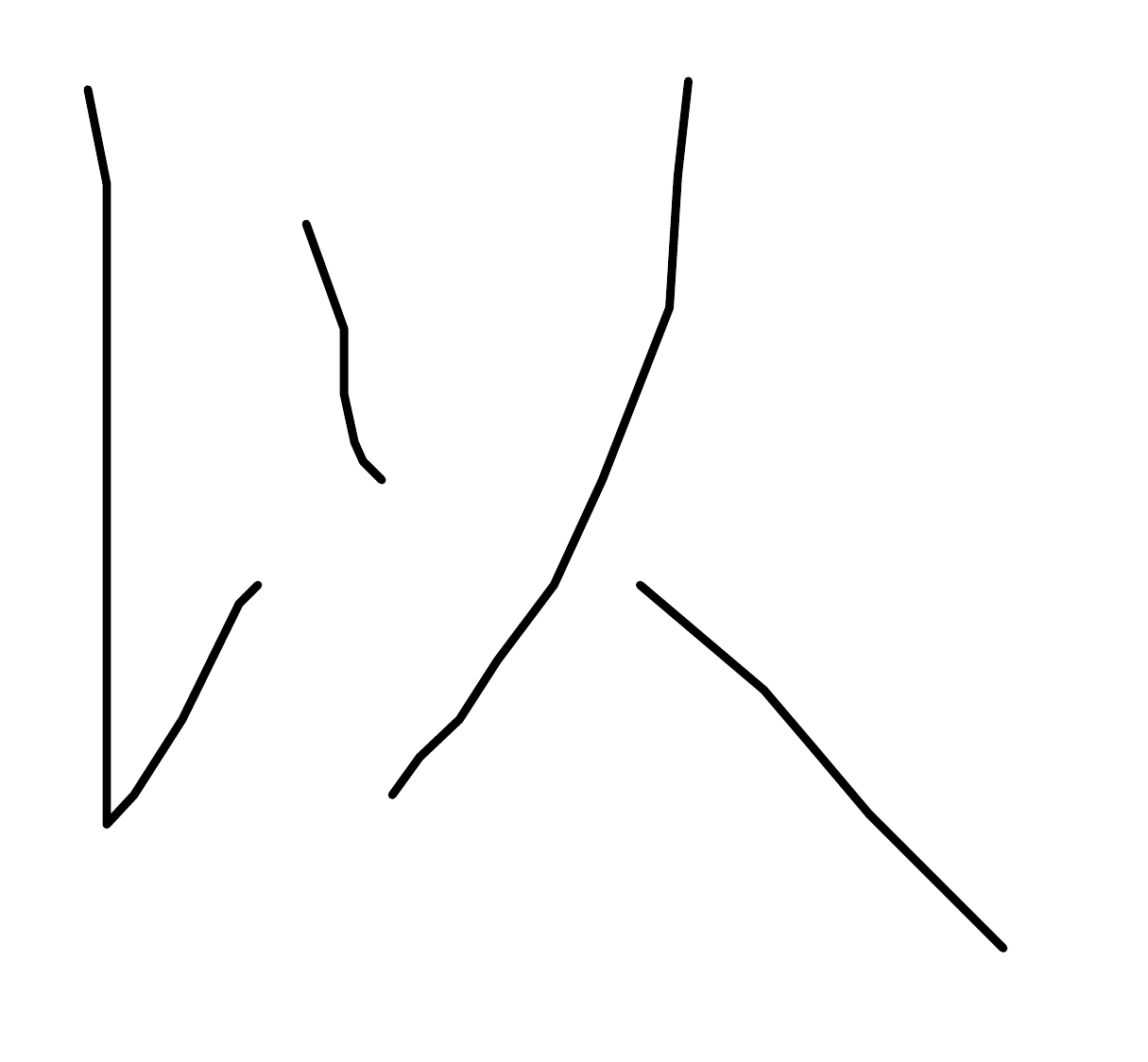}}\hspace{5truemm}
\fbox{\includegraphics[width=0.7in,height=0.7in]{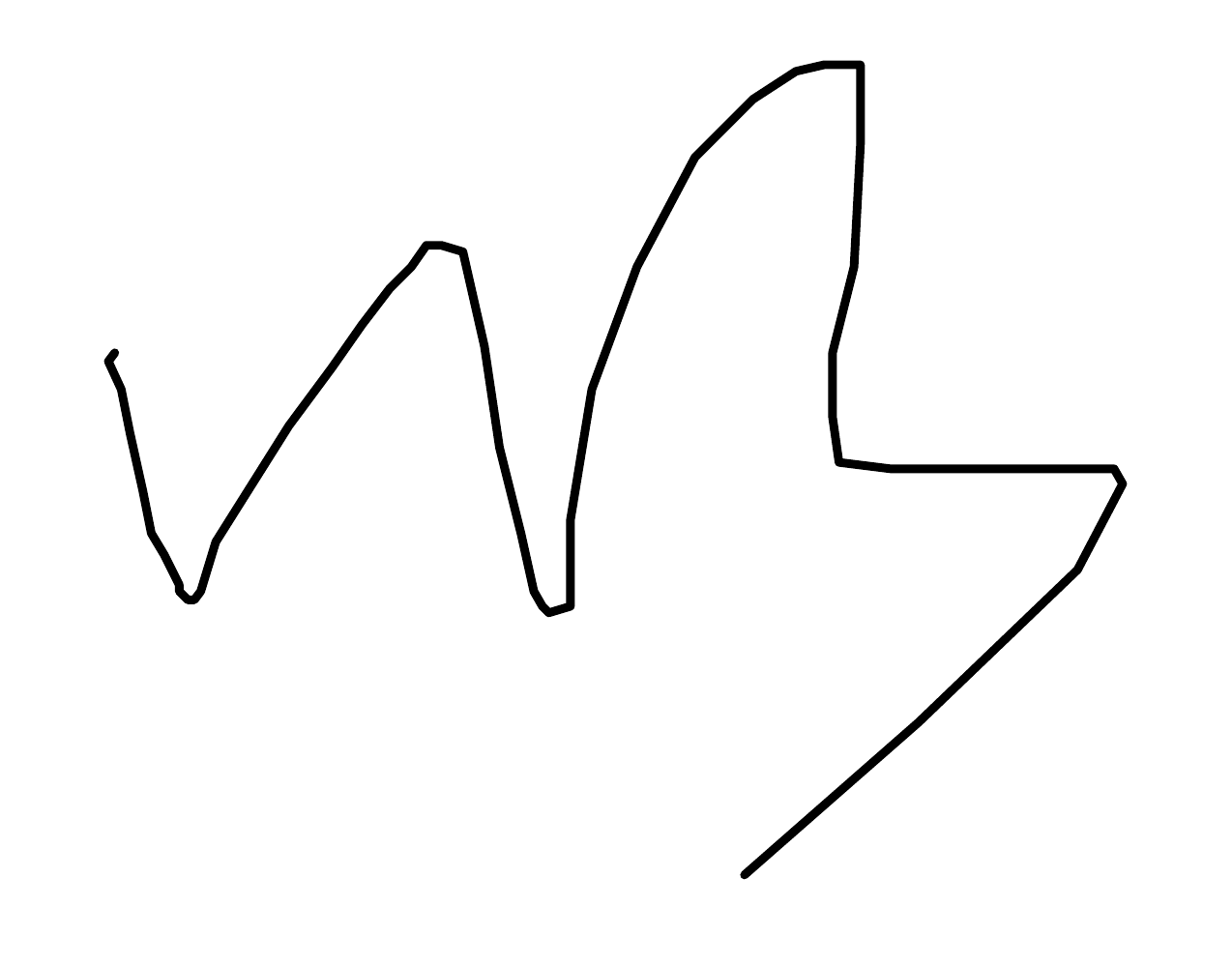}}\hspace{5truemm}
\fbox{\includegraphics[width=0.7in,height=0.7in]{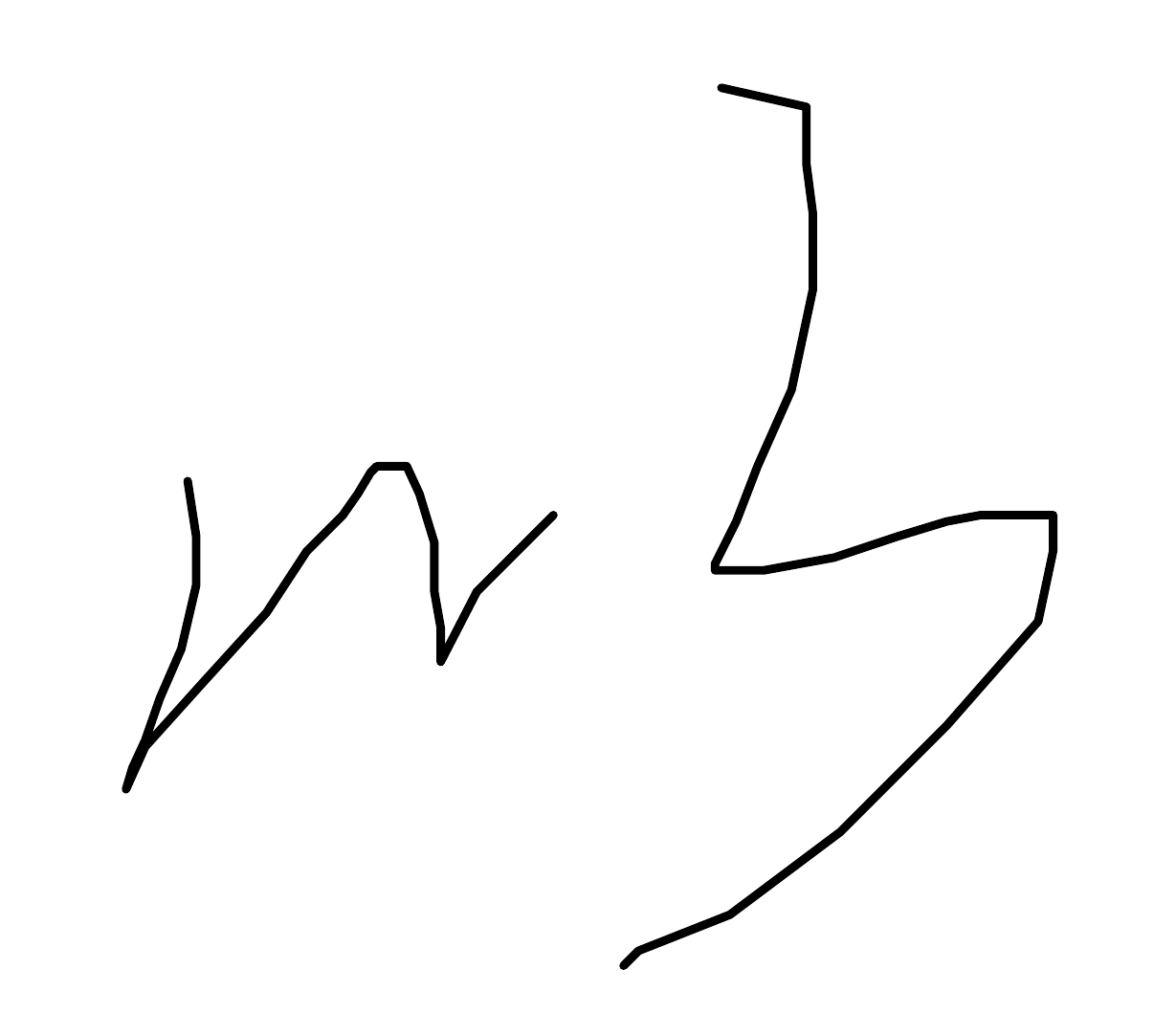}}
\vspace{-3truemm}
\caption{\sl Variations of U+4EE5  (\protect\zhs{以})  }
\label{4ee5print}
\end{figure}

\begin{figure}[!h]
\vspace{-2truemm}
\centering
\fbox{\includegraphics[width=0.7in,height=0.7in]{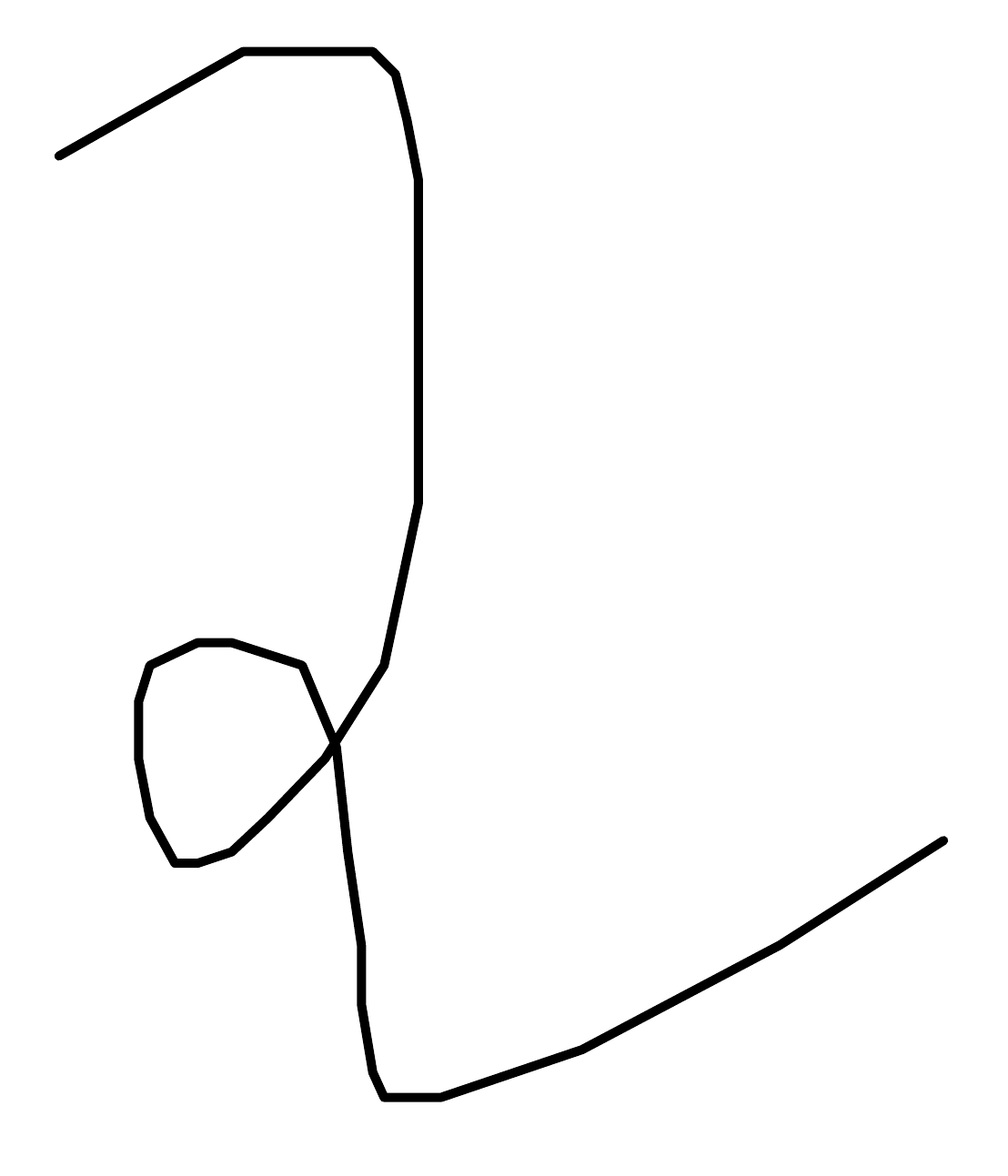}}\hspace{5truemm}
\fbox{\includegraphics[width=0.7in,height=0.7in]{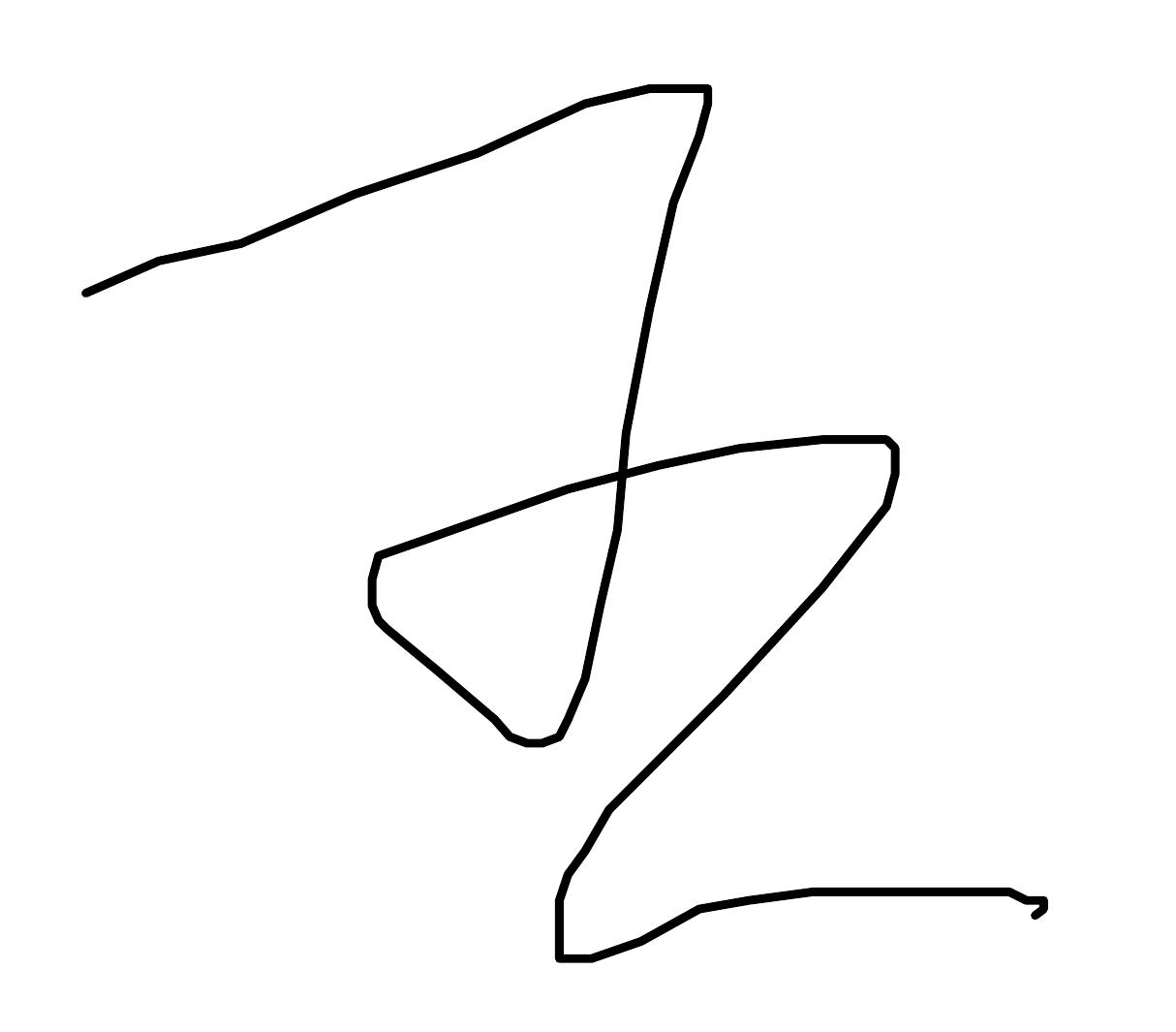}}
\vspace{-3truemm}
\caption{\sl Similar Shapes of U+738B (\protect\zhs{王}) and U+4E94 (\protect\zhs{五})  }
\label{jadevsfive}
\vspace{-2truemm}
\end{figure}

The fact that, in daily life, users often write quickly and
unconstrained can lead to cursive variations that have a very
dissimilar appearance. Conversely, sometimes it also leads to
confusability between different characters. Just like in English,
where a ``v'' becomes a ``u'' shape when writing quickly, two
different Chinese cursive characters could end up looking very similar
to each other. Figs. ~\ref{duprint}--\ref{jadevsfive} show some of the
concrete examples we observe in our data. Note that it is especially
important to have enough training material to distinguish cursive
variations such as in Fig.~\ref{jadevsfive}.

\begin{table}[t]
\vspace{-7truemm}
\renewcommand{\arraystretch}{0.9}
\caption{Model Hyper-Parameters}
\label{tab:architecture_hp}
\centering
\begin{tabular}{|c|c|}
\hline
Hyper-parameters & Values\\
\hline
expansion factor $t$  & [2, 4, 6, 8]\\
nb. block sequence & [3, 4, 5] \\
max. sequence len. & [2, 4, 6, 8, 10]\\
first conv. channels & [16, 24, 32, 48]\\
last conv. channels & [128, 256]\\
fully conn. layer units  & [0, 128, 256]\\
\hline
\end{tabular}
\vspace{-5truemm}
\end{table}

\section{Experimental Results}
\label{sec:experiments}
\vspace{-1truemm}
\subsection{Model optimization}
Several MobileNetV2 hyper-parameters require fine-tuning:
Table~\ref{tab:architecture_hp} reflects a family of parameters that
yields good accuracy. We use the same expansion factor for all the
bottleneck blocks. The number of bottleneck block sequences is
selected to decrease the spatial resolution of the last feature map
and hence decrease the model disk footprint. The number of ouput
channels of the last convolution layer and the number of units of the
hidden fully-connected layer are also chosen to limit the model
footprint. In addition, we experimented with replacing the
fully-connected layer with a global average pooling layer.  The
sequence length of each bottleneck block is uniformly sampled in the
range [1, max. sequence len.], providing some diversity on the
sequence length within a given model. Finally, the numbers of channels
of the bottleneck blocks are chosen from the following list [32, 48,
64, 96, 128, 256].

We ensure that the number of channels from one block sequence to the
next one never decreases.  Starting by the first bottleneck block, the
number of output channels is randomly sampled from the first 3 entries
of the list. The list is then updated by discarding any values
strictly smaller than the sampled value. Next, the number of channels
for the next layer is chosen from the updated list. This sampling and
list updating process is carried on for all the bottleneck blocks, and
the maximum number of channels is restricted to be 256.

We use stochastic gradient descent with Nesterov momentum as
optimizer. We found that a learning rate between 0.01 and 0.1 and a
batch of size between 250 and 1000 are reasonable choices.
Additionally, we supplemented actual observations with suitable
elastic deformations as advocated in \cite{meier, simard}.  Finally,
we tried to include attention modules at the output of the bottleneck
blocks~ \cite{selfAttention}. However, it did not provide any
immediate accuracy improvement; further investigation is probably
warranted but left for future work.

For every training scenario detailed below for various datasets,
we launched 50 model configuration trials by randomly sampling
hyper-parameters values from Table~\ref{tab:architecture_hp}, and
reported the accuracy and disk footprint of the best model. On the
average, we observed an accuracy delta between the 10 best
configurations of 0.6\%. In order to showcase the diversity of the
model architecture explored, Table~\ref{tab:architecture_type1} and
Table~\ref{tab:architecture_type2} show the configurations of the two
best models, respectively named type 1 and type 2. It is worth noting
that despite being quite different, both models provide a similar
level of accuracy. It suggests that there is not a single optimal
configuration, but instead a family of configurations that yields high
accuracy. Interestingly, the last hidden layer of both these models
have 128 units, leading to relatively compact models with a disk
footprint of 19MB for the type 1 model and of 17MB for the type 2
model. Similar configurations with 256 units in their last hidden
layers exhibit a disk footprint above 30MB.

\begin{table}[!htb]
\vspace{-5truemm}
\renewcommand{\arraystretch}{0.9}
\caption{Architecture Type 1}
\label{tab:architecture_type1}
\centering
\begin{tabular}{|c|c|c|c|c|c|}
\hline
Input & Layer & $t$ & $c$ & $n$ & $s$\\
\hline
48 $\times$ 48 $\times$ 1 & conv2D & - & 16 & 1 & 1\\
48 $\times$ 48 $\times$ 16 & bottleneck & 8 & 64 & 4 & 2\\
24 $\times$ 24 $\times$ 64 & bottleneck & 8 & 64 & 4 & 2\\
12 $\times$ 12 $\times$ 64 & bottleneck & 8 & 64 & 4 & 2\\
6 $\times$ 6 $\times$ 64 & bottleneck & 8 & 96 & 2 & 2\\
3 $\times$ 3 $\times$ 96 & conv2D & - & 128 & 1 & 1\\
3 $\times$ 3 $\times$ 128 & dense & - & 128 & 1 & -\\
128 & dense & - & 30K & 1 & -\\
\hline
\end{tabular}
\end{table}

\begin{table}[!htb]
\vspace{-9truemm}
\renewcommand{\arraystretch}{0.9}
\caption{Architecture Type 2}
\label{tab:architecture_type2}
\centering
\begin{tabular}{|c|c|c|c|c|c|}
\hline
Input & Layer & $t$ & $c$ & $n$ & $s$\\
\hline
48 $\times$ 48 $\times$ 1 & conv2D & - & 24 & 1 & 1\\
48 $\times$ 48 $\times$ 24 & bottleneck & 4 & 32 & 7 & 2\\
24 $\times$ 24 $\times$ 32 & bottleneck & 4 & 64 & 8 & 2\\
12 $\times$ 12 $\times$ 64 & bottleneck & 4 & 96 & 4 & 2\\
6 $\times$ 6 $\times$ 96 & conv2D & - & 128 & 1 & 1\\
6 $\times$ 6 $\times$ 128 & global avg. pooling  & - & 128 & 1 & 1\\
128 & dense & - & 30K & 1 & -\\
\hline
\end{tabular}
\vspace{-5truemm}
\end{table}

\subsection{Results on CASIA dataset}
\label{subsec:results}
Whereas the ultimate goal is to scale up to GB18030-like coverage, it
is informative to start by evaluating our MobileNetV2 implementation
on a benchmark task such as CASIA \cite{casia}. While covering
relatively few characters, this task has the merit to have been
well-studied in the literature, as reported in, e.g.,
\cite{benchmarks2011} and \cite{benchmarks2013}. For comparison, we
replicated the same setup based on the datasets CASIA-OLHWDB,
DB1.0-1.2 \cite{casia,benchmarks2011}, split in training and testing
datasets, yielding about one million training exemplars.

Table~\ref{tab:casiaBaseline} shows the results obtained using the
architecture of Fig.~1, where the abbreviation ``Hz--1'' refers to the
H\`anz\`i level-1 inventory (3,755 characters), and ``CR($n$)''
denotes top-$n$ character recognition accuracy. Note that we are
listing top-4 accuracy in addition to commonly reported top-1 and
top-10 accuracies: because our user interface was designed to show 4
character candidates, top-4 accuracy is an important predictor of user
experience in our system.

\begin{table}[t]
\vspace{-2truemm}
\renewcommand{\arraystretch}{1.0}
\caption{\sl Results on CASIA On-line Database, 3,755 Characters. Standard
  Training, $\hbox{\sl Associated Model Size} = 11 \hbox{\sl MB}$.}
\label{tab:casiaBaseline}
\vspace{1truemm}
\centering
\ninept
\begin{tabular}{|c|c||c|c|c|}
\hline
Inventory & Training & CR(1) & CR(4) & CR(10) \\
\hline
Hz--1      & CASIA    & 95.1\%  & 98.9\%  & 99.5\% \\
\hline
\end{tabular}
\vspace{-4truemm}
\end{table}

The figures in Table~\ref{tab:casiaBaseline} compare with on-line
results in \cite{benchmarks2011} and \cite{benchmarks2013} averaging
roughly 93\% for top-1 and 98\% for top-10 accuracy. Thus, our top-1
and top-10 accuracies are in line with the literature. Furthermore,
our top-4 accuracy is very satisfactory, and perhaps even more
importantly, obtained with a model size (11 MB) on the smaller end of
the spectrum among comparable systems in \cite{benchmarks2011} and
\cite{benchmarks2013}.

The system in Table~\ref{tab:casiaBaseline} is trained only on CASIA
data, and does not include any other training data. We were also
interested in folding in additional training data collected in-house
on a variety of devices. As detailed in Section~\ref{sec:dataset},
this data covers a larger variety of styles and comprises a lot more
training instances per character. Table~\ref{tab:casia_augmented}
reports the results observed, on the same test set with a
3,755-character inventory.

\begin{table}[t]
\renewcommand{\arraystretch}{1.0}
\caption{\sl Results on CASIA On-line Database, 3,755
  Characters. Augmented Training, $\hbox{\sl Associated Model Size} = 13 \hbox{\sl MB}$.}
\label{tab:casia_augmented}
\vspace{1truemm}
\centering
\ninept
\begin{tabular}{|c|c||c|c|c|}
\hline
Inventory & Training & CR(1) & CR(4) & CR(10) \\
\hline
Hz--1     & Augmented & 96.8\% & 99.4\%  & 99.7\% \\
\hline
\end{tabular}
\vspace{-2truemm}
\end{table}

The resulting top-1 and top-4 accuracies are 1.7\% and 0.5\% higher,
respectively, when compared to the system in
Table~\ref{tab:casiaBaseline}. Such relatively modest improvement
suggests that the various character styles appearing in the test set
were already well covered in the CASIA training set. But it also
indicates that folding in additional styles has no deleterious effect
on the model, which in turn supports looking at learning curves to
estimate how the error scales with data size
\cite{nipsCortesJSVD93}. As is well-known, on a log-log plot the
``steepness'' of such learning curves empirically conveys how quickly
a model can learn from adding more training samples \cite{scaling}.

Finally, Table~\ref{tab:fullTraining_casia} reports results on the
exact same test set for the full system, for which the number of
recognizable characters increases from 3,755 to approximately
30,000. Compared to Table~\ref{tab:casia_augmented}, accuracy
drops---which was to be expected, since the vastly increased coverage
creates additional confusability (for example, between \zhs{二} and
``Z'' as mentioned earlier). Still, the drop remains rather limited,
which is encouraging. In fact, a comparison between Tables
\ref{tab:casia_augmented} and \ref{tab:fullTraining_casia} shows that
multiplying coverage by a factor of 8 entails far less than 8 times
more errors, or 8 times more storage. Instead, the increase in both
number of errors and model size is confined to manageable
values. Thus, building a high-accuracy Chinese character recognition
that covers 30,000 characters, instead of only 3,755, is possible and
practical.

\begin{table}[t]
\vspace{-2truemm}
\renewcommand{\arraystretch}{1.0}
\caption{\sl Results on CASIA On-line Database, 30,000 Characters.}
\label{tab:fullTraining_casia}
\vspace{1truemm}
\centering
\ninept
\begin{tabular}{|c||c|c|c|c|}
\hline
Inventory & CR(1) & CR(4) & CR(10)  & Model Size \\
\hline
30K          & 96.6\%   & 99.3\%  & 99.6\%     & 19\hbox{\rm MB} \\
\hline
\end{tabular}
\vspace{-4truemm}
\end{table}

\vspace{-3truemm}
\subsection{Results on in-house dataset}
\label{subsubsec:inhouse_dataset}
To get an idea of how the full system performs across the entire set
of 30,000 characters, we also evaluated the accuracy of the
MobileNetV2 architecture of type 1
(Table~\ref{tab:architecture_type1}) on a number of different test
sets comprising all supported characters written in various styles. As
baseline, we opted for the LeNet architecture, as it has been commonly
used in previous handwriting recognition experiments on the MNIST task
(see, e.g., \cite{ciresan}, \cite{meier}). The LeNet model is composed
of 3 convolution layers, each followed by max pooling, and a
fully-connected hidden layer with 128 units, similarly to the
MobileNetV2 model. This model has a similar disk footprint as the
MobileNetV2 model. Table~\ref{tab:fullTraining_inHouse} lists the
average results for both models on the in-house test sets.  Despite
both models having roughly the same number of parameters, the
mobileNetV2 model top-1 accuracy is 4.6\% higher than the LeNet
model. This highlights the progress made by the community in designing
CNN architectures in recent years.

\begin{table}[t]
\renewcommand{\arraystretch}{1.0}
\caption{\sl Average Results on Multiple In-House Test Sets Comprising All
  Writing Styles, 30,000 Characters.}
\label{tab:fullTraining_inHouse}
\vspace{1truemm}
\centering
\ninept
\begin{tabular}{|c||c|c|c|c|}
\hline
Model & CR(1) & CR(4) & CR(10) & Model Size\\
\hline
mobileNetV2  & 97.2\%   & 99.6\%  & 99.8\% &  19\hbox{\rm MB} \\
LeNet    & 92.6\%   & 98.4\%  & 99.2\% & 15\hbox{\rm MB} \\
\hline
\end{tabular}
\vspace{-2truemm}
\end{table}

Finally, even though the results in
Tables~\ref{tab:fullTraining_casia}--\ref{tab:fullTraining_inHouse}
are not directly comparable since they were obtained on different test
sets, they show that top-1 and top-4 accuracies are in the same
ballpark across the entire inventory of characters. This was to be
expected given the largely balanced training regimen.

\section{Conclusion}
\label{sec:discussion}
We have discussed the unique challenges involved in recognizing
Chinese handwritten characters spanning a large inventory of
approximately 30,000 characters, while simultaneously achieving
real-time performance and minimizing disk memory footprint. Particular
attention has to be paid to data collection conditions,
representativeness of writing styles, and training regimen.  Our
experimental observations show that, with proper care, a high-accuracy
handwriting recognition system is practical on mobile
devices. Furthermore, accuracy only degrades slowly as the supported
inventory increases, as long as training data of sufficient quality is
available in sufficient quantity. So, how well would we expect to
handle even larger inventories?  The total set of CJK characters in
Unicode is currently around 75,000 \cite{CJKUnicode}, and the
ideographic rapporteur group (IRG) keeps on suggesting new additions
from a variety of sources. Admittedly, these new characters will be
rare (e.g., used in historical names or poetry). Nevertheless, they
are of high interest to every person dealing with them.

Learning curves \cite{nipsCortesJSVD93, scaling} obtained by varying
the amount of training data allow for extrapolation of asymptotic
values, regarding both what our accuracy would look like with more
training data, and how it would change with more characters to
recognize. Given the 8-times larger inventory and corresponding (less
than) 0.2\% drop in accuracy between Table~\ref{tab:casia_augmented}
and Table~\ref{tab:fullTraining_casia}, we can extrapolate that with
an inventory of 100,000 characters and a corresponding increase in
training data, it would be realistic to achieve top-1 accuracies
around 96\%, and top-10 accuracies around 99\% (with the same type of
architecture). These figures support recognition feasibility, even on
mobile devices. This bodes well for the recognition of even larger
sets of Chinese characters in the future.

\end{document}